\documentclass[runningheads]{llncs}
\usepackage{graphicx}
\usepackage{comment}
\usepackage{url}
\usepackage{amsmath}
\usepackage{amssymb}
\usepackage{booktabs}
\usepackage{amsfonts}
\usepackage{multirow}
\usepackage{balance}
\usepackage{tikz}
\usepackage{subfigure}
\begin{document}

\title{{GBDF}: Gender Balanced DeepFake Dataset Towards Fair DeepFake Detection}

\titlerunning{Gender Balanced DeepFake Dataset Towards Fair DeepFake Detection}
%
\author{Aakash Varma Nadimpalli \and
Ajita Rattani\thanks{Corresponding author}}
\authorrunning{Nadimpalli and Rattani}
%
\institute{School of Computing \\
Wichita State University, USA \\
\email{axnadimpalli@shockers.wichita.edu}, 
\email{ajita.rattani@wichita.edu}}
\maketitle              
\begin{abstract}
Facial forgery by deepfakes has raised severe societal concerns. Several solutions have been proposed by the vision community to effectively combat the misinformation on the internet via automated deepfake detection systems. Recent studies have demonstrated that facial analysis-based deep learning models can discriminate based on protected attributes. For the commercial adoption and massive roll-out of the deepfake detection technology, it is vital to evaluate and understand the fairness (the absence of any prejudice or favoritism) of deepfake detectors across demographic variations such as gender and race. As the performance differential of deepfake detectors between demographic sub-groups would impact millions of people of the deprived sub-group.
This paper aims to evaluate the fairness of the deepfake detectors across males and females. However, existing deepfake datasets are not annotated with demographic labels to facilitate fairness analysis.  To this aim, we manually annotated existing popular deepfake datasets with gender labels and evaluated the performance differential of current deepfake detectors across gender. Our analysis on the gender-labeled version of the datasets suggests (a) current deepfake datasets have skewed distribution across gender, and (b) commonly adopted deepfake detectors obtain unequal performance across gender with mostly males outperforming females. Finally, 
we contributed a gender-balanced and annotated deepfake dataset, GBDF, to mitigate the performance differential and to promote research and development towards fairness-aware deep fake detectors.
The GBDF dataset is publicly available at: \url{https://github.com/aakash4305/GBDF}




\keywords{DeepFakes  \and Fairness and Bias in AI \and Facial Analysis.}
\end{abstract}
\section{Introduction}

With the advances in deep generative models, synthetic media have become so realistic that they are often indiscernible from authentic content for human eyes. However, synthetic media generation techniques used by malicious users to deceive pose a severe societal and political threat. 
In this context, Deepfakes - facial forgery technique that depicts human subjects with altered identities or malicious actions using various deep fake generation techniques- has been flagged as a top AI threat~\cite{Nguyen2019DeepLF,citron,Tolosana2020deepfakesAB,9157215,Verdoliva2020MediaFA}. Deep fakes have been used to commit fraud, falsify evidence, manipulate public debates, and destabilize political processes~\cite{cellan-jones_2019,Tolosana2020deepfakesAB}. 

\definecolor{rosso}{RGB}{220,57,18}
\definecolor{giallo}{RGB}{255,153,0}
\definecolor{blu}{RGB}{102,140,217}
\definecolor{verde}{RGB}{16,150,24}
\definecolor{viola}{RGB}{153,0,153}
\makeatletter
\tikzstyle{chart}=[
    legend label/.style={font={\scriptsize},anchor=west,align=left},
    legend box/.style={rectangle, draw, minimum size=5pt},
    axis/.style={black,semithick,->},
    axis label/.style={anchor=east,font={\tiny}},
]
\tikzstyle{bar chart}=[
    chart,
    bar width/.code={
        \pgfmathparse{##1/2}
        \global\let\bar@w\pgfmathresult
    },
    bar/.style={very thick, draw=white},
    bar label/.style={font={\bf\small},anchor=north},
    bar value/.style={font={\footnotesize}},
    bar width=.75,
]
\tikzstyle{pie chart}=[
    chart,
    slice/.style={line cap=round, line join=round, very thick,draw=white},
    pie title/.style={font={\bf}},
    slice type/.style 2 args={
        ##1/.style={fill=##2},
        values of ##1/.style={}
    }
]
\pgfdeclarelayer{background}
\pgfdeclarelayer{foreground}
\pgfsetlayers{background,main,foreground}
\newcommand{\pie}[3][]{
    \begin{scope}[#1]
    \pgfmathsetmacro{\curA}{90}
    \pgfmathsetmacro{\r}{1}
    \def\c{(0,0)}
    \node[pie title] at (90:1.3) {#2};
    \foreach \v/\s in{#3}{
        \pgfmathsetmacro{\deltaA}{\v/100*360}
        \pgfmathsetmacro{\nextA}{\curA + \deltaA}
        \pgfmathsetmacro{\midA}{(\curA+\nextA)/2}
        \path[slice,\s] \c
            -- +(\curA:\r)
            arc (\curA:\nextA:\r)
            -- cycle;
        \pgfmathsetmacro{\d}{max((\deltaA * -(.5/50) + 1) , .5)}
        \begin{pgfonlayer}{foreground}
        \path \c -- node[pos=\d,pie values,values of \s]{$\v\%$} +(\midA:\r);
        \end{pgfonlayer}
        \global\let\curA\nextA
    }
    \end{scope}
}
\newcommand{\legend}[2][]{
    \begin{scope}[#1]
    \path
        \foreach \n/\s in {#2}
            {
                  ++(0,-10pt) node[\s,legend box] {} +(5pt,0) node[legend label] {\n}
            }
    ;
    \end{scope}
}

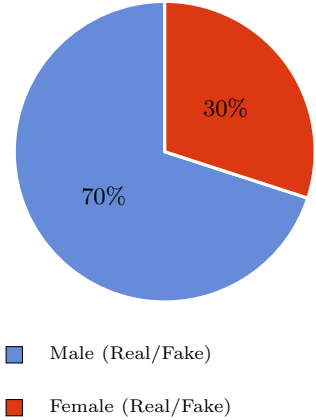
\begin{figure*}[htbp]
\centering
\begin{subfigure}
    \centering
\begin{tikzpicture}
[
    pie chart,
    slice type={comet}{blu},
    slice type={legno}{rosso},
    slice type={coltello}{giallo},
    slice type={sedia}{viola},
    slice type={caffe}{verde},
    pie values/.style={font={\small}},
    scale=2
]
    \pie{FaceForensics++ Distribution}{23.56/comet,41.31/legno,35.12/coltello}
%
    \legend[shift={(-1cm,-1cm)}]{{Male (Real/Fake)}/comet, {Female (Real/Fake)}/legno, {Irregular (Fake)}/coltello}
\end{tikzpicture}
\end{subfigure}
    \hfill
\begin{subfigure}
    \centering
\begin{tikzpicture}
[
    pie chart,
    slice type={comet}{blu},
    slice type={legno}{rosso},
    slice type={coltello}{giallo},
    slice type={sedia}{viola},
    slice type={caffe}{verde},
    pie values/.style={font={\small}},
    scale=2
]

    \pie{CelebDF Distribution}{70/comet,30/legno}
%
    \legend[shift={(-1cm,-1cm)}]{{Male (Real/Fake)}/comet, {Female (Real/Fake)}/legno}

\node at (0,-2.1) {};
\end{tikzpicture}
\end{subfigure}
    \hfill

\caption{Illustration of the distribution of videos in Face Forensics++ and Celeb-DF Dataset across gender. The percentage distribution of videos belonging to males (real/fake), females (real/fake) and those classified as irregular swaps is shown. 
} 
\label{fig_one}
\end{figure*}

To mitigate the risk posed by deep fakes, the vision community has developed a series of effective deep fake detection methods~\cite{Nguyen2019DeepLF,citron,Tolosana2020deepfakesAB} trained on large-scale deepfake datasets. The popular deep fake detection methods include convolutional neural networks~(CNN) for detecting visual artifacts~\cite{Matern2019ExploitingVA} and blending boundaries~\cite{9157215}, mouth movement analysis~\cite{9578910} and behavioral biometrics~\cite{9360904}. The popular publicly available deep fake datasets include Celeb-DF~\cite{9156368}, FaceForensics++~\cite{9010912}, DeeperForensics-$1.0$~\cite{jiang2020deeperforensics1} and DFDC~\cite{https://doi.org/10.48550/arxiv.2006.07397} for research and development in this field. 

Such efforts have been translated into creating \textbf{real-world impact} with Microsoft's release of Video Authenticator\footnote{\url{https://blogs.microsoft.com/on-the-issues/2020/09/01/disinformation-deepfakes-newsguard-video-authenticator/}}, an automated tool trained on the publicly available FaceForensics++ deepfake dataset, for detection of artificial manipulation in images and videos. Further, Facebook\footnote{\url{https://www.wired.com/story/facebook-removes-accounts-ai-generated-photos/}} has been advancing its methods to detect and ban AI-generated profiles, along with strengthening its policy on deepfakes and synthetic media. Recently, the Coalition for Content Provenance and Authenticity (C2PA) has teamed up with Intel, and Adobe to develop new standards targeted at combating the proliferation of deepfakes\footnote{\url{https://c2pa.org/post/release_1_pr/}}.

While significant advances have been made towards accurate deepfake detection, very little is discussed on the fairness of these deepfake detectors across protected attributes (demographic variations) such as gender and race. Fairness is defined as the absence of any prejudice or favoritism towards an individual or a group based on their inherent or acquired characteristics~\cite{10.1109/TIFS.2021.3135750,9356331}. For the \textit{massive commercial roll-out} of deep fake detection technology, it is vital to examine the bias and fairness of this technology across demographics. This is to avoid any real-world consequences from a biased and flawed system toward a particular sub-group. \textit{As in the common operating scenario, the social media data across gender and race would be audited at the mass level for authenticity via an automated deepfake detection system. Even the small performance differential of deepfake detectors across demographic sub-groups would impact millions of people belonging to the deprived sub-group.} 

This draws attention to fairness and bias in AI-based facial analytics where unintended consequences from biased systems call for a thorough examination of the datasets and models~\cite{9356331,10.1109/TIFS.2021.3135750,karkkainenfairface,buolamwini2018gender,albiero2020does}. Most of the published research in this domain suggests low performance for women, and dark-skinned people for facial attribute-based classification systems such as gender and age~\cite{buolamwini2018gender,karkkainenfairface,singh2021anatomizing,Nadimpalli2021HarnessingUD}, and face recognition~\cite{10.1109/TIFS.2021.3135750,albiero2020does}. 
As biased datasets produce biased models, many of the efforts have been focused on developing gender and race-balanced datasets for various facial-analysis based applications. FairFace~\cite{karkkainenfairface}, a gender and race balanced facial attribute dataset, RFW~\cite{Wang2019RacialFI}, a racially balanced face recognition dataset, and a gender-balanced dataset developed from existing facial recognition datasets~\cite{albiero2020does} are some of the examples.

This paper aims to examine the \textbf{fairness} of deepfake detectors across gender. 
 However, current deepfake detection datasets are not annotated with demographic labels to facilitate the examination of bias. To this aim, deepfake datasets namely FaceForensics++,and Celeb-DF are \textit{manually annotated} with gender labels. The fairness of popular deepfake detectors is evaluated on these datasets across gender. On manual annotation, we found that the gender distribution of the popular deepfake datasets is skewed. The large number of deepfakes in Faceforensics++ are irregular (in conformance with~\cite{Trinh2021AnEO})- when a person's face is swapped with the face of another gender or race. This result in the loss of gender-specific information in the fake content. The popular Celeb-DF dataset distribution is heavily skewed towards males ($70\%$).
 
 Figure~\ref{fig_one} shows the distribution of videos across gender for the popular FaceForensics++~\cite{9010912} and Celeb-DF~\cite{9156368} deepfake datasets. The deepfake detectors evaluated on these skewed datasets along with irregular swaps mostly obtain lower performance for females over males. 
  Finally, we introduced a gender-balanced and annotated deepfake dataset, GBDF, developed from FaceForensics++, Celeb-DF, and DeeperForensics-1.0 and consisting of $10,000$ videos. This balanced dataset aims to mitigate the performance differential of deepfake detectors due to existing gender unbalanced training sets along with irregular swaps. The dataset information is available to the vision community to promote further research and development in this field. Note that according to ISO/IEC 22116~\cite{ISO_gender}, the term ``sex”, understood as ``the state of being male or female” would be more appropriate instead of ``gender” in the context of this study. However, in consistency with the existing studies~\cite{buolamwini2018gender,albiero2020does}, the term gender is used in this paper.
 To the best of our knowledge, the only study in~\cite{Trinh2021AnEO} evaluates the bias of three popular CNN-based deepfake detectors trained on Faceforensics++ across gender and race. The test bed was created using UTKFace and RFW datasets and the deepfakes were generated using the Face X-ray model. The authors reported performance differences for dark-skinned people and emphasized the importance of benchmark representation and auditing for increased demographic transparency.

The main \textbf{contributions} of the paper are as follows:
\begin{enumerate}

\item Gender label annotation of the popular deepfake datasets namely, FaceForensics++ and Celeb-DF to facilitate analysis of the dataset distribution across gender and the presence of irregular swaps.

\item Evaluation of the fairness of popular deepfake detection algorithms varying in size, architecture, and the methodology, trained and tested on gender annotated versions of the existing datasets. 

\item Development of publicly available gender-balanced and annotated deepfake dataset, GBDF, from FaceForensics++ (FF++), Celeb-DF, and Deeper Forensics-1.0 consisting of $10,000$ live and fake videos generated using different identity and expression swapping deepfake generation techniques. 

\item Cross-comparison of the performance differential of deepfake detectors trained on existing and our gender-balanced GBDF training set, across males and females. 

\end{enumerate}

This paper is organized as follows: Section 2 discuss the related work on deepfake detectors and gendered differences in facial analytics. Section 3 discusses the development of the GBDF dataset. Deepfake detection algorithms used in this study are discussed in Section 4. Evaluation metrics used for fairness analysis are discussed in Section 5. Results and discussion is detailed in Section 6. Conclusion and future research directions are discussed in Section 7. 

\section{Related Work}

\subsection{Deepfake Detection}

In this section, we will discuss the existing countermeasure proposed for deep fake detection. Most of the existing methods are CNN-based classification baselines trained for deep fake detection~\cite{He2016DeepRL,8099678,Nguyen2019MultitaskLF,9717407}. 

In ~\cite{Li_2019_CVPR_Workshops}, Li and Lyu used VGG16, ResNet50, ResNet101, and ResNet152 based CNNs for the detection of the presence of artifacts from the facial regions and the surrounding areas for deep fake detection. Afchar et al.~\cite{8630761} proposed two different CNN architectures composed of only a few layers in order to focus on the mesoscopic properties of the images: (a) a CNN comprised of 4 convolutional layers followed by a fully-connected layer (Meso-4), and (b) a modification of Meso-4 using a variant of the Inception module named MesoInception-4. In~\cite{9010912}, an exhaustive analysis of different CNN-based deep fake detection methods by Rosslet et al. suggested efficacy of XceptionNet when evaluated on FaceForensics++. In~\cite{9157215}, a face X-ray model has been proposed to detect forgery by detecting the blending boundary of a forged image using a two-class CNN model trained end-to-end.

Apart from the aforementioned CNN-based deep fake detection methods, spatial temporal information using Long Short-term Memory~(LSTM) networks~\cite{10.1145/3369412.3395070},  facial  and behavioral biometrics (i.e., facial expression, head, and body movement), and lipforensics~\cite{9578910} have been used for deep fake detection~\cite{Dong2020IdentityDrivenDD,9360904,Agarwal2019ProtectingWL,9717407}. In~\cite{9578910}, LipForensics that targets high-level semantic irregularities in mouth movements common in many generated deepfake videos, is used for deepfake detection. 
Studies have also been proposed for improving the performance of deepfake detectors across datasets and deep fake generation methods using techniques such as reinforcement learning~\cite{nadimpalli2022improving} and fine-grained multi-attention network~\cite{9577592}.
Readers are referred to the published survey in~\cite{Tolosana2020deepfakesAB},~\cite{Nguyen2019DeepLF} for detailed information on deep fake detection methods.

\subsection{Gendered Differences in Facial Analytics}
There is consensus in the published literature that face analytics-based computer vision applications obtain lower accuracy for females, who often have both a higher false match and a higher false non-match rate over males~\cite{buolamwini2018gender,albiero2020does,9356331,10.1109/TIFS.2021.3135750,karkkainenfairface}. Examination of the fairness of the gender classification systems using commercial SDKs and deep learning-based CNNs suggest lower accuracy rates for females consistently~\cite{9356331,buolamwini2018gender}. 
$2019$ Face Recognition Vendor test documents lower female accuracy rates across a broad range of algorithms and datasets\footnote{~\url{https://www.nist.gov/system/files/documents/2019/11/20/frvt_report_2019_11_19_0.pdf}}.  Similarly, lower accuracy rates for females have been obtained for various in-house deep learning-based face recognition systems~\cite{10.1109/TIFS.2021.3135750,albiero2020does,singh2021anatomizing}. The cause and effect analysis suggests gendered hairstyles resulting in facial occlusion, make-up, and inherent lower variability between different female faces over males to be the factors contributing to lower performance for females~\cite{albiero2020does,10.1109/TIFS.2021.3135750}. The demographic balanced datasets have been proven to mitigate the performance differential of different facial analysis based applications across demographics~\cite{albiero2020does,9356331,karkkainenfairface}.

\section{GBDF: Gender Balanced DeepFake Dataset}

The GBDF dataset is created using FF++($c23$ version), Celeb-DF, Deeper Forensics-$1.0$ and consist of $10,000$ videos with $5000$ each for males and females. 

 \noindent The FaceForensics++~\cite{9010912} (FF++) is an automated benchmark for facial manipulation detection. It consists of several manipulated videos created using two different generation techniques: Identity Swapping (FaceSwap, FaceSwap-Kowalski, FaceShifter, Deep Fakes) and Expression swapping (Face2Face and NeuralTextures). The Celeb-DF~\cite{9156368} deep fake forensic dataset include $590$ genuine videos from $59$ celebrities as well as $5639$ deep fake videos. Celeb-DF, in contrast to other datasets, has essentially no splicing borders, color mismatch, and inconsistencies in face orientation, among other evident deep fake visual artifacts. The deep fake videos in Celeb-DF are created using an encoder-decoder style model which results in better visual quality. The DeeperForensics-$1.0$~\cite{jiang2020deeperforensics1} is one of the largest deep fake datasets used for face forgery detection. It consists of $60,000$ videos that have around $17.6$ million frames with substantial real-world perturbations. The dataset contains videos of $100$ consented actors with $35$ different perturbations. The real to fake videos ratio is $5$:$1$ and the fake videos are generated by an end-to-end face-swapping framework. 

\noindent \textbf{Gender Label Annotation}. As none of these existing deepfake datasets contain demographic information, we manually annotated ground truth gender labels for these datasets. To do so, we annotated each subject with the perceived gender {male, female}. Two graduate annotators were selected for the task of gender label annotation. For each subject, the annotators were presented with an average of $150$ frames at various times in the video, which displayed the subject at different light angles and poses. The gender label was assigned to each video based on the consensus between the annotators. 
With the annotated gender labels, we evaluated the percentage of videos belonging to males, and females and those being irregular face-swaps. Recall that an irregular swap is defined as a swap where a person's face is swapped onto another person's face of a different gender. All the three datasets provided the IDs for pairs of swaps for all the manipulation methods, With the help of the available IDs which are unique for all the identities, we were able to segregate gender labels as well as irregular swaps.
FaceForensics++ has $35.12\%$ of irregular deepfakes. Irregular deepfakes were not found in Celeb-DF. DeeperForensics-1.0 dataset has negligible number of irregular swaps. To remain ethnically aware and to maintain demographic information, irregular swaps from FaceForensics++ and deeperforensics-$1.0$ datasets are not included in the GBDF dataset.
 
The gender annotated version of the live and deepfake videos (excluding irregular swaps) from these deepfakes datasets are merged to create GBDF dataset. Deepfakes in the GBDF dataset are created using different Identity Swapping (i.e., FaceSwap, FaceSwap-Kowalski, FaceShifter, Encoder-decoder style and End-to-end Face Swapping techniques) and Expression swapping 
(i.e., Face2Face and NeuralTextures) deepfake generation techniques. 
The majority of the videos in GBDF are from Caucasians.
The ratio of real to fake videos in the GBDF dataset is $1:4$. The GBDF is further divided into gender-balanced and subject independent training and testing subsets in the ratio of $70:30$. 
Figure~\ref{fig_two} illustrates the comparison of deepfake videos among existing Deepfake datasets and our GBDF. The number of videos in GBDF is higher than many of the existing deepfake datasets shown on the x axis.
The GBDF dataset is publicly available at: \url{https://github.com/aakash4305/GBDF}

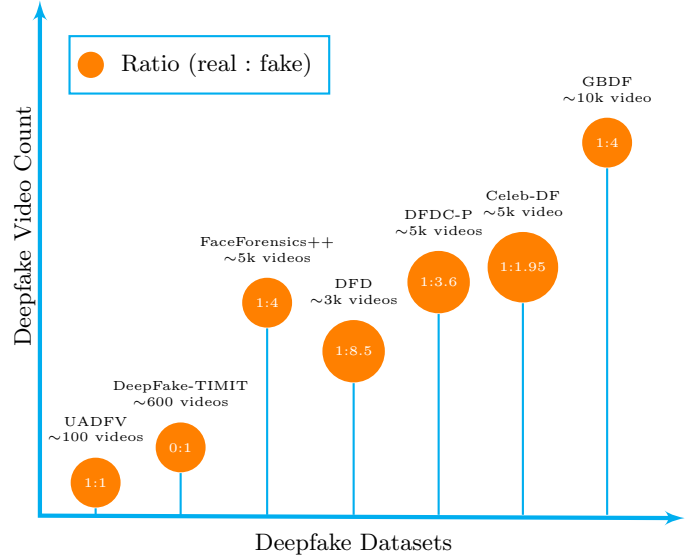
\begin{figure*}[htbp]
\centering

\begin{tikzpicture}
\usetikzlibrary{fit, positioning}
\usetikzlibrary{shapes,arrows}
 \tikzstyle{green_circle} = [ shape=circle, text=white, very thick, draw= orange!100!black, top color= orange!100!black,bottom color= orange!100!black]
  \tikzstyle{green_line} = [ cyan!100!black]
\node (v2) at (-0.1708,7.2325) {};
\node (v1) at (-0.1709,0.3789) {};
\node (v3) at (8.3999,0.3464) {};
\draw [latex'-latex', ultra thick, green_line] (v2.center) -- (v1.center) -- (v3.center);
\node  (label_ratio) at (2.1799,6.3766) {Ratio (real : fake)};
\node (label_ratio_c) [ left = 1.5cm of label_ratio.center, green_circle] {};
\node[draw, green_line, thick, fit={(label_ratio) (label_ratio_c)}] {};
     
\node [rotate=90] at (-0.3708,4.2325) {Deepfake Video Count};
%
 
\draw [thick, green_line] (0.57,0.39) node (v15) {} -- (0.57,0.4767) node (v4) {};
\draw  [thick, green_line](1.70,0.39) node (v14) {} -- (1.70,0.9453) node (v5) {};
\draw  [thick, green_line](2.85,0.39) node (v13) {} -- (2.85,2.8705) node (v6) {};
\draw  [thick, green_line](4,0.39) node (v12) {} -- (4,2.1414) node (v7) {};
\draw  [thick, green_line](5.13,0.39) node (v11) {} -- (5.13,3.0612) node (v8) {};
\draw  [thick, green_line](6.25,0.39) node (v10) {} -- (6.25,3.205) node (v9) {};
\draw  [thick, green_line](7.37,0.39) node (v90) {} -- (7.37,5) node (v99) {};
\node (v20) [above= 0.01cm of v4.center, green_circle, font=\tiny, align=center] {1:1};
\node  (v21)  [above= 0.01cm of v5.center, green_circle, font=\tiny, align=center] {0:1};
\node  (v22)  [above= 0.01cm of v6.center, green_circle, font=\tiny, align=center] {1:4};
\node  (v23)  [above= 0.01cm of v7.center, green_circle, font=\tiny, align=center] {1:8.5};
\node  (v24)  [above= 0.01cm of v8.center, green_circle, font=\tiny, align=center] {1:3.6};
\node  (v25)  [above= 0.01cm of v9.center, green_circle, font=\tiny, align=center] {1:1.95};
\node  (v26)  [above= 0.01cm of v99.center, green_circle, font=\tiny, align=center] {1:4};
\node [above=0.1cm of v20, font=\tiny, align=center] {UADFV \\ $\sim$100 videos};
\node [above=0.1cm of v21, font=\tiny, align=center] {DeepFake-TIMIT \\ $\sim$600 videos};
\node [above=0.1cm of v22, font=\tiny, align=center] {FaceForensics++ \\$\sim$5k videos};
\node [above=0.1cm of v23, font=\tiny, align=center] {DFD \\$\sim$3k videos};
\node [above=0.1cm of v24, font=\tiny, align=center] {DFDC-P\\$\sim$5k videos};
\node [above=0.1cm of v25, font=\tiny, align=center] {Celeb-DF\\$\sim$5k video};
\node [above=0.1cm of v26, font=\tiny, align=center] {GBDF\\$\sim$10k video};
\node at (4,0) {Deepfake Datasets};
\end{tikzpicture}
\caption{Illustration of the number of videos in different deepfake datasets along with our proposed GBDF dataset. The figure contains information about the real to fake ratio of videos in the datasets along with deepfake video count. The no. of videos in GBDF is higher than many of the existing deepfake datasets shown on the x-axis.
} 
\label{fig_two}
\end{figure*}

\section{Deepfake Detection Algorithms Used}
We investigated fairness of popular deepfake detection models of various sizes, architectures and the underlying concept, across males and females. Specifically, we trained MesoInception4\footnote{\url{https://github.com/HongguLiu/MesoNet-Pytorch}}, XceptionNet\footnote{\url{https://github.com/i3p9/deepfake-detection-with-xception}}, EfficientNet V2-L\footnote{\url{https://github.com/d-li14/efficientnetv2.pytorch}}, LipForensics\footnote{\url{https://github.com/ahaliassos/LipForensics}} and CNN-LSTM~\footnote{\url{https://github.com/oidelima/Deepfake-Detection}} based deepfake detectors.

These models are trained on the popular FF++ dataset($c23$ version) and our proposed GBDF training set. 
We used the sampling approach described in~\cite{9010912} to choose $270$ frames per video for training and $150$ frames per video for validation and testing of most of the models. The face images were detected and aligned using MTCNN~\cite{7553523} algorithm. MTCNN utilizes a cascaded CNN based framework for joint face detection and alignment. The images are then resized to $256\times256$ for both training and evaluation.  
 
For all the CNN-based models, we used a batch-normalization layer followed by the last fully connected layer of size $1024$ and the final output layer for deep fake classification. The CNN models were trained using an Adam optimizer with an initial learning rate of $0.001$ and a weight decay of 1e6.  
For CNN-LSTM model, we chose EfficientNet V2-L as the backbone CNN model due to its superior performance. The CNN network's output of $2048$ feature vector is fed into the LSTM layer for deepfake detection. 
For LipForensics model, following authors implementation in~\cite{9578910}, the network receives $25$ grayscale, aligned mouth crops of size $88\times 88$ as input for each video. The input is passed through pretrained ResNet-18 (pretrained for lipreading task with an initial 3-D convolutional layer) to obtain output embedding sensitive to mouth motion analysis. A multiscale temporal convolutional network (MS-TCN) was finetuned to detect fake videos based on semantically high-level anomalies in mouth motion, which was also pretrained for lipreading task. All the models were trained on $4$ RTX $5000$Ti GPUs with a batch size of $64$.

\section{Evaluation Metrics}
Following the standard evaluation metrics adopted for deepfake detectors, we used partial AUC (pAUC) (at $10\%$ False Positive Rate~(FPR)) and Equal Error Rate~(EER) for the evaluation of performance differences across males and females. 
Further, as deepfake detection is a binary classification task, we have also analyzed binary classification metrics for fairness evaluation across males and females. Similar to the bias evaluation study on gender classification by Buolamwini et al.~\cite{buolamwini2018gender}, we follow the evaluation precedent established by the National Institute of Standards and Technology (NIST) and assessed the overall classification accuracy (ACC), along with the true positive rate~(TPR), and false-positive rate~(FPR) for males and females.

\section{Results and Analysis}
In this section, we examine the fairness of the deepfake detectors, discussed in section $4$, across males and females on FF++, Celeb-DF, GBDF, and an external DFDC-P~\cite{https://doi.org/10.48550/arxiv.2006.07397} test sets.  All the \textbf{evaluation metrics} (from section $5$) are reported in the range $[0, 1]$.

\subsection{Performance differential of deepfake detectors on FF++ test set}


\begin{table*}[]
\caption {Evaluation of the DeepFake Detectors Across Males and Females when trained on FF++, GBDF and tested on \textbf{FF++}. The metrics used are AUC, pAUC and EER. The performance differential (P.D) is also calculated as the absolute difference between EER of males and females.}
\label{Table1}
\begin{center}
\scalebox{0.78}{
\begin{tabular}{|c|c|ccc|ccc|ccc|c|}
\hline
\multirow{2}{*}{\textbf{Models}} & \multirow{2}{*}{\textbf{Training Dataset}} & \multicolumn{3}{c|}{\textbf{Overall}}                                                      & \multicolumn{3}{c|}{\textbf{Male}}                                                         & \multicolumn{3}{c|}{\textbf{Female}}                                                       & \multirow{2}{*}{\textbf{P.D}$\downarrow$} \\ \cline{3-11}
                                 &                                            & \multicolumn{1}{c|}{\textbf{AUC}}   & \multicolumn{1}{c|}{\textbf{pAUC}}  & \textbf{EER}   & \multicolumn{1}{c|}{\textbf{AUC}}   & \multicolumn{1}{c|}{\textbf{pAUC}}  & \textbf{EER}   & \multicolumn{1}{c|}{\textbf{AUC}}   & \multicolumn{1}{c|}{\textbf{pAUC}}  & \textbf{EER}   &                               \\ \hline
EfficientNet V2-L               & \multirow{5}{*}{FaceForensics++}                      & \multicolumn{1}{c|}{\textbf{0.991}} & \multicolumn{1}{c|}{\textbf{0.979}} & \textbf{0.024} & \multicolumn{1}{c|}{\textbf{0.995}} & \multicolumn{1}{c|}{\textbf{0.986}} & \textbf{0.019} & \multicolumn{1}{c|}{0.987}          & \multicolumn{1}{c|}{0.972}          & 0.029          & 0.010                         \\ \cline{1-1} \cline{3-12} 
XceptionNet                     &                                            & \multicolumn{1}{c|}{0.985}          & \multicolumn{1}{c|}{0.969}          & 0.037          & \multicolumn{1}{c|}{0.987}          & \multicolumn{1}{c|}{0.975}          & 0.029          & \multicolumn{1}{c|}{0.983}          & \multicolumn{1}{c|}{0.963}          & 0.045          & 0.016                         \\ \cline{1-1} \cline{3-12} 
MesoInception-4                  &                                            & \multicolumn{1}{c|}{0.857}          & \multicolumn{1}{c|}{0.832}          & 0.229          & \multicolumn{1}{c|}{0.863}          & \multicolumn{1}{c|}{0.837}          & 0.221          & \multicolumn{1}{c|}{0.851}          & \multicolumn{1}{c|}{0.827}          & 0.237          & 0.016                         \\ \cline{1-1} \cline{3-12} 
CNN-LSTM                         &                                            & \multicolumn{1}{c|}{0.987}          & \multicolumn{1}{c|}{0.972}          & 0.032          & \multicolumn{1}{c|}{0.991}          & \multicolumn{1}{c|}{0.979}          & 0.024          & \multicolumn{1}{c|}{0.983}          & \multicolumn{1}{c|}{0.967}          & 0.039          & 0.015                         \\ \cline{1-1} \cline{3-12} 
LipForensics                  &                                            & \multicolumn{1}{c|}{0.990}          & \multicolumn{1}{c|}{0.977}          & 0.027          & \multicolumn{1}{c|}{0.987}          & \multicolumn{1}{c|}{0.975}          & 0.031          & \multicolumn{1}{c|}{\textbf{0.993}} & \multicolumn{1}{c|}{\textbf{0.979}} & \textbf{0.023} & \textbf{0.008}                \\ \hline
EfficientNet V2-L               & \multirow{5}{*}{GBDF}                      & \multicolumn{1}{c|}{0.925}          & \multicolumn{1}{c|}{0.902}          & 0.136          & \multicolumn{1}{c|}{0.935}          & \multicolumn{1}{c|}{0.917}          & 0.121          & \multicolumn{1}{c|}{0.915}          & \multicolumn{1}{c|}{0.888}          & 0.140          & 0.019                         \\ \cline{1-1} \cline{3-12} 
XceptionNet                     &                                            & \multicolumn{1}{c|}{0.906}          & \multicolumn{1}{c|}{0.886}          & 0.176          & \multicolumn{1}{c|}{0.912}          & \multicolumn{1}{c|}{0.892}          & 0.172          & \multicolumn{1}{c|}{0.899}          & \multicolumn{1}{c|}{0.880}          & 0.180          & \textbf{0.008}                \\ \cline{1-1} \cline{3-12} 
MesoInception-4                  &                                            & \multicolumn{1}{c|}{0.806}          & \multicolumn{1}{c|}{0.785}          & 0.264          & \multicolumn{1}{c|}{0.813}          & \multicolumn{1}{c|}{0.794}          & 0.259          & \multicolumn{1}{c|}{0.799}          & \multicolumn{1}{c|}{0.775}          & 0.269          & 0.010                         \\ \cline{1-1} \cline{3-12} 
CNN-LSTM                         &                                            & \multicolumn{1}{c|}{0.918}          & \multicolumn{1}{c|}{0.897}          & 0.141          & \multicolumn{1}{c|}{0.910}          & \multicolumn{1}{c|}{0.890}          & 0.150          & \multicolumn{1}{c|}{0.926}          & \multicolumn{1}{c|}{0.904}          & 0.132          & 0.018                         \\ \cline{1-1} \cline{3-12} 
LipForensics                  &                                            & \multicolumn{1}{c|}{\textbf{0.932}} & \multicolumn{1}{c|}{\textbf{0.917}} & \textbf{0.122} & \multicolumn{1}{c|}{\textbf{0.937}} & \multicolumn{1}{c|}{\textbf{0.921}} & \textbf{0.117} & \multicolumn{1}{c|}{\textbf{0.928}} & \multicolumn{1}{c|}{\textbf{0.913}} & \textbf{0.128} & 0.011                         \\ \hline
\end{tabular}}
\end{center}
\end{table*}
Table~\ref{Table1} shows the performance of the deepfake detectors across males and females when trained on FF++, GBDF, and tested on FF++. Similarly, Table~\ref{Table2} shows the corresponding ACC, TPR, and FPR values of these models. 
The top performance results are highlighted in bold across various evaluation datasets. EfficientNet V2-L obtained the best results with an overall AUC of $0.991$, EER of $0.024$, and ACC of $0.975$ when trained and tested on FF++. 

When trained on FF++, the overall difference in the performance is $0.009$ and $0.010$ in terms of pAUC and EER, respectively, across males and females. Males outperformed females for the majority of the models despite having a lower percentage than females in FF++ training set. \textit{The reason is $35.12\%$ of the videos in FF++ are irregular deepfakes, it is not certain which gender-group-related features are dominant in irregular facial swaps}.
The overall difference in ACC, TPR, and FPR is $0.006$,$0.0036$, and $0.020$, respectively, across males and females~(see Table~\ref{Table2}). The least performance differential is obtained by LipForensics model when trained and tested on FF++. 

\begin{table*}[]
\caption {ACC, TPR and FPR of the DeepFake Detectors Across Males and Females when trained on FF++, GBDF and tested on FF++. When trained on GBDF, the drop in the performance of the models is due to domain shift. The GBDF dataset consist of higher number of deepfake generation techniques over FF+.} 
\label{Table2}
\begin{center}
\scalebox{0.88}{
\begin{tabular}{|c|c|ccc|ccc|ccc|}
\hline
\multirow{2}{*}{Models} & \multirow{2}{*}{Training Datasets} & \multicolumn{3}{c|}{Overall}                                                               & \multicolumn{3}{c|}{Male}                                                                  & \multicolumn{3}{c|}{Female}                                                                \\ \cline{3-11} 
                        &                                    & \multicolumn{1}{c|}{ACC}            & \multicolumn{1}{c|}{TPR}            & FPR            & \multicolumn{1}{c|}{ACC}            & \multicolumn{1}{c|}{TPR}            & FPR            & \multicolumn{1}{c|}{ACC}            & \multicolumn{1}{c|}{TPR}            & FPR            \\ \hline
EfficientNet V2-L      & \multirow{5}{*}{FaceForensics++}              & \multicolumn{1}{c|}{\textbf{0.975}} & \multicolumn{1}{c|}{\textbf{0.952}} & \textbf{0.091} & \multicolumn{1}{c|}{\textbf{0.979}} & \multicolumn{1}{c|}{\textbf{0.955}} & \textbf{0.058} & \multicolumn{1}{c|}{0.971}          & \multicolumn{1}{c|}{0.949}          & 0.119          \\ \cline{1-1} \cline{3-11} 
XceptionNet            &                                    & \multicolumn{1}{c|}{0.969}          & \multicolumn{1}{c|}{0.942}          & 0.128          & \multicolumn{1}{c|}{0.971}          & \multicolumn{1}{c|}{0.947}          & 0.109          & \multicolumn{1}{c|}{0.967}          & \multicolumn{1}{c|}{0.937}          & 0.139          \\ \cline{1-1} \cline{3-11} 
MesoInception-4         &                                    & \multicolumn{1}{c|}{0.825}          & \multicolumn{1}{c|}{0.805}          & 0.256          & \multicolumn{1}{c|}{0.834}          & \multicolumn{1}{c|}{0.813}          & 0.245          & \multicolumn{1}{c|}{0.816}          & \multicolumn{1}{c|}{0.797}          & 0.267          \\ \cline{1-1} \cline{3-11} 
CNN-LSTM                &                                    & \multicolumn{1}{c|}{0.971}          & \multicolumn{1}{c|}{0.945}          & 0.115          & \multicolumn{1}{c|}{0.976}          & \multicolumn{1}{c|}{0.954}          & 0.093          & \multicolumn{1}{c|}{0.966}          & \multicolumn{1}{c|}{0.936}          & 0.137          \\ \cline{1-1} \cline{3-11} 
LipForensics          &                                    & \multicolumn{1}{c|}{0.972}          & \multicolumn{1}{c|}{0.948}          & 0.115          & \multicolumn{1}{c|}{0.967}          & \multicolumn{1}{c|}{0.941}          & 0.143          & \multicolumn{1}{c|}{\textbf{0.978}} & \multicolumn{1}{c|}{\textbf{0.955}} & \textbf{0.086} \\ \hline
EfficientNet V2-L      & \multirow{5}{*}{GBDF}              & \multicolumn{1}{c|}{0.903}          & \multicolumn{1}{c|}{0.887}          & 0.182          & \multicolumn{1}{c|}{0.912}          & \multicolumn{1}{c|}{0.895}          & 0.175          & \multicolumn{1}{c|}{0.892}          & \multicolumn{1}{c|}{0.879}          & 0.187          \\ \cline{1-1} \cline{3-11} 
XceptionNet            &                                    & \multicolumn{1}{c|}{0.888}          & \multicolumn{1}{c|}{0.869}          & 0.189          & \multicolumn{1}{c|}{0.897}          & \multicolumn{1}{c|}{0.876}          & 0.181          & \multicolumn{1}{c|}{0.879}          & \multicolumn{1}{c|}{0.862}          & 0.195          \\ \cline{1-1} \cline{3-11} 
MesoInception-4         &                                    & \multicolumn{1}{c|}{0.783}          & \multicolumn{1}{c|}{0.769}          & 0.284          & \multicolumn{1}{c|}{0.794}          & \multicolumn{1}{c|}{0.778}          & 0.276          & \multicolumn{1}{c|}{0.772}          & \multicolumn{1}{c|}{0.760}          & 0.292          \\ \cline{1-1} \cline{3-11} 
CNN-LSTM                &                                    & \multicolumn{1}{c|}{0.896}          & \multicolumn{1}{c|}{0.875}          & 0.185          & \multicolumn{1}{c|}{0.887}          & \multicolumn{1}{c|}{0.861}          & 0.189          & \multicolumn{1}{c|}{0.905}          & \multicolumn{1}{c|}{0.889}          & 0.178          \\ \cline{1-1} \cline{3-11} 
LipForensics          &                                    & \multicolumn{1}{c|}{\textbf{0.912}} & \multicolumn{1}{c|}{\textbf{0.896}} & \textbf{0.176} & \multicolumn{1}{c|}{\textbf{0.919}} & \multicolumn{1}{c|}{\textbf{0.901}} & \textbf{0.169} & \multicolumn{1}{c|}{\textbf{0.905}} & \multicolumn{1}{c|}{\textbf{0.891}} & \textbf{0.183} \\ \hline
\end{tabular}}
\end{center}
\end{table*}

When trained on GBDF, the overall difference in the performance is $0.010$ and $0.006$ in terms of pAUC and EER, respectively, across males and females.
The overall difference in ACC, TPR, and FPR was reduced to $0.011$, $0.006$ and $0.009$, respectively, across males and females~(see Table~\ref{Table2}).  XceptionNet model obtained the least performance differential when trained on GBDF and tested on FF++.

Therefore, the overall difference in EER and FPR was reduced to $0.04$ and $0.011$, respectively, when using GBDF over FF++ as the training set. Using GBDF as the training set, the highest bias mitigation is observed for XceptionNet with the EER difference reduced from $0.016$ to $0.008$ across gender. Most of the detectors obtained lower error rates when trained on FF++. This is obvious as the test bed is also FF++. The performance of most of the models dropped when trained using GBDF due to domain shift i.e., the data distribution change between the training (GBDF) and testing set (FF++). This is due to change in the image quality of real videos and deep fakes due to advances in sensor technology and the deep fake generation techniques. The GBDF dataset has an additional number of deepfake generation techniques (based on encoder-decoder style and the end-to-end face swapping framework) over FF++.

\subsection{Performance differential of deepfake detectors on Celeb-DF test set}

\begin{table*}[]
\caption {Evaluation of the DeepFake Detectors Across Males and Females when trained on FF++, GBDF and tested on \textbf{Celeb-DF}. The metrics used are AUC, pAUC and EER. The performance differential (P.D) is also calculated as the absolute difference between EER of males and females.}
\label{Table3}
\begin{center}
\scalebox{0.78}{
\begin{tabular}{|c|c|ccc|ccc|ccc|c|}
\hline
\multirow{2}{*}{\textbf{Models}} & \multirow{2}{*}{\textbf{Training Dataset}} & \multicolumn{3}{c|}{\textbf{Overall}}                                                      & \multicolumn{3}{c|}{\textbf{Male}}                                                         & \multicolumn{3}{c|}{\textbf{Female}}                                                       & \multirow{2}{*}{\textbf{P.D}$\downarrow$} \\ \cline{3-11}
                                 &                                            & \multicolumn{1}{c|}{\textbf{AUC}}   & \multicolumn{1}{c|}{\textbf{pAUC}}  & \textbf{EER}   & \multicolumn{1}{c|}{\textbf{AUC}}   & \multicolumn{1}{c|}{\textbf{pAUC}}  & \textbf{EER}   & \multicolumn{1}{c|}{\textbf{AUC}}   & \multicolumn{1}{c|}{\textbf{pAUC}}  & \textbf{EER}   &                               \\ \hline
EfficientNet V2-L               & \multirow{5}{*}{FaceForensics++}                      & \multicolumn{1}{c|}{0.658}          & \multicolumn{1}{c|}{0.635}          & 0.379          & \multicolumn{1}{c|}{0.667}          & \multicolumn{1}{c|}{0.645}          & 0.372          & \multicolumn{1}{c|}{0.649}          & \multicolumn{1}{c|}{0.625}          & 0.386          & 0.014                         \\ \cline{1-1} \cline{3-12} 
XceptionNet                     &                                            & \multicolumn{1}{c|}{0.651}          & \multicolumn{1}{c|}{0.629}          & 0.383          & \multicolumn{1}{c|}{0.657}          & \multicolumn{1}{c|}{0.634}          & 0.379          & \multicolumn{1}{c|}{0.645}          & \multicolumn{1}{c|}{0.623}          & 0.390          & \textbf{0.011}                \\ \cline{1-1} \cline{3-12} 
MesoInception-4                  &                                            & \multicolumn{1}{c|}{0.544}          & \multicolumn{1}{c|}{0.519}          & 0.459          & \multicolumn{1}{c|}{0.558}          & \multicolumn{1}{c|}{0.528}          & 0.442          & \multicolumn{1}{c|}{0.530}          & \multicolumn{1}{c|}{0.510}          & 0.476          & 0.034                         \\ \cline{1-1} \cline{3-12} 
CNN-LSTM                         &                                            & \multicolumn{1}{c|}{0.675}          & \multicolumn{1}{c|}{0.656}          & 0.359          & \multicolumn{1}{c|}{0.686}          & \multicolumn{1}{c|}{0.662}          & 0.348          & \multicolumn{1}{c|}{0.664}          & \multicolumn{1}{c|}{0.650}          & 0.370          & 0.022                         \\ \cline{1-1} \cline{3-12} 
LipForensics                  &                                            & \multicolumn{1}{c|}{\textbf{0.821}} & \multicolumn{1}{c|}{\textbf{0.795}} & \textbf{0.254} & \multicolumn{1}{c|}{\textbf{0.829}} & \multicolumn{1}{c|}{\textbf{0.805}} & \textbf{0.242} & \multicolumn{1}{c|}{\textbf{0.813}} & \multicolumn{1}{c|}{\textbf{0.785}} & \textbf{0.266} & 0.024                         \\ \hline
EfficientNet V2-L               & \multirow{5}{*}{GBDF}                      & \multicolumn{1}{c|}{0.861}          & \multicolumn{1}{c|}{0.844}          & 0.235          & \multicolumn{1}{c|}{0.869}          & \multicolumn{1}{c|}{0.853}          & 0.228          & \multicolumn{1}{c|}{0.853}          & \multicolumn{1}{c|}{0.835}          & 0.242          & 0.014                         \\ \cline{1-1} \cline{3-12} 
XceptionNet                     &                                            & \multicolumn{1}{c|}{0.864}          & \multicolumn{1}{c|}{0.847}          & 0.233          & \multicolumn{1}{c|}{0.872}          & \multicolumn{1}{c|}{0.855}          & 0.226          & \multicolumn{1}{c|}{0.856}          & \multicolumn{1}{c|}{0.839}          & 0.240          & 0.014                         \\ \cline{1-1} \cline{3-12} 
MesoInception-4                  &                                            & \multicolumn{1}{c|}{0.742}          & \multicolumn{1}{c|}{0.725}          & 0.298          & \multicolumn{1}{c|}{0.755}          & \multicolumn{1}{c|}{0.735}          & 0.292          & \multicolumn{1}{c|}{0.730}          & \multicolumn{1}{c|}{0.715}          & 0.305          & 0.013                         \\ \cline{1-1} \cline{3-12} 
CNN-LSTM                         &                                            & \multicolumn{1}{c|}{0.887}          & \multicolumn{1}{c|}{0.869}          & 0.215          & \multicolumn{1}{c|}{0.898}          & \multicolumn{1}{c|}{0.875}          & 0.209          & \multicolumn{1}{c|}{0.876}          & \multicolumn{1}{c|}{0.863}          & 0.221          & \textbf{0.012}                \\ \cline{1-1} \cline{3-12} 
LipForensics                  &                                            & \multicolumn{1}{c|}{\textbf{0.908}} & \multicolumn{1}{c|}{\textbf{0.885}} & \textbf{0.175} & \multicolumn{1}{c|}{\textbf{0.917}} & \multicolumn{1}{c|}{\textbf{0.896}} & \textbf{0.163} & \multicolumn{1}{c|}{\textbf{0.900}} & \multicolumn{1}{c|}{\textbf{0.874}} & \textbf{0.187} & 0.024\\ \hline
\end{tabular}}
\end{center}
\end{table*}

Table~\ref{Table3} shows the performance differential of the deepfake detectors when trained on FF++, GBDF, and tested on Celeb-DF. Similarly, Table~\ref{Table4} shows the corresponding ACC, TPR, and FPR values for these models. The top performance results are highlighted in bold across various evaluation datasets. The LipForensics model obtained the best results with an overall AUC of $0.908$, EER of $0.175$, and ACC of $0.889$ when trained on GBDF and tested on Celeb-DF. 

When trained on FF++, the overall difference in the performance is $0.0162$ and $0.021$ in terms of pAUC and EER, respectively, across males and females.  The overall difference in ACC, TPR and FPR is $0.019$,$0.02$ and $0.021$, respectively, across males and females (see Table~\ref{Table4}). The least performance differential is obtained by XceptionNet when trained on FF++ and tested on Celeb-DF.

When trained on GBDF, the overall difference in the performance is $0.017$ and $0.015$ in terms of pAUC and EER, respectively, across males and females. The overall difference in ACC, TPR and FPR is $0.018$,$0.01$ and $0.018$, respectively, across males and females~(see Table~\ref{Table4}). The least performance differential is obtained by CNN-LSTM when trained on GBDF and tested on Celeb-DF.

Therefore, the difference in AUC, EER, TPR, and FPR is reduced to $0.001$, $0.006$, $0.01$, and $0.003$, respectively, when using GBDF as a training set over FF++. Using GBDF, the highest bias mitigation is observed for MesoInceptionNet-4 model with the EER difference reduced from $0.034$ to $0.013$ across gender. 
The overall performance of all the models increased when trained on GBDF over FF++ because of the presence of higher number of deepfake generation techniques. \textbf{It is worth noting that the training and testing subset of GBDF and Celeb-DF, respectively, has no subject overlap}.
This experiment points out the \textbf{merit} of using a demographically balanced dataset for deepfake detection.  




\begin{table*}[]
\caption {ACC, TPR and FPR of the DeepFake Detectors Across Males and Females when trained on FF++, GBDF and tested on Celeb-DF.}
\label{Table4}
\begin{center}
\scalebox{0.88}{
\begin{tabular}{|c|c|ccc|ccc|ccc|}
\hline
\multirow{2}{*}{Models} & \multirow{2}{*}{Training Datasets} & \multicolumn{3}{c|}{Overall}                                                               & \multicolumn{3}{c|}{Male}                                                                  & \multicolumn{3}{c|}{Female}                                                                \\ \cline{3-11} 
                        &                                    & \multicolumn{1}{c|}{ACC}            & \multicolumn{1}{c|}{TPR}            & FPR            & \multicolumn{1}{c|}{ACC}            & \multicolumn{1}{c|}{TPR}            & FPR            & \multicolumn{1}{c|}{ACC}            & \multicolumn{1}{c|}{TPR}            & FPR            \\ \hline
EfficientNet V2-L      & \multirow{5}{*}{FaceForensics++}              & \multicolumn{1}{c|}{0.637}          & \multicolumn{1}{c|}{0.604}          & 0.385          & \multicolumn{1}{c|}{0.650}          & \multicolumn{1}{c|}{0.614}          & 0.372          & \multicolumn{1}{c|}{0.626}          & \multicolumn{1}{c|}{0.594}          & 0.398          \\ \cline{1-1} \cline{3-11} 
XceptionNet            &                                    & \multicolumn{1}{c|}{0.629}          & \multicolumn{1}{c|}{0.602}          & 0.395          & \multicolumn{1}{c|}{0.635}          & \multicolumn{1}{c|}{0.609}          & 0.383          & \multicolumn{1}{c|}{0.623}          & \multicolumn{1}{c|}{0.595}          & 0.402          \\ \cline{1-1} \cline{3-11} 
MesoInception-4         &                                    & \multicolumn{1}{c|}{0.525}          & \multicolumn{1}{c|}{0.502}          & 0.437          & \multicolumn{1}{c|}{0.534}          & \multicolumn{1}{c|}{0.518}          & 0.422          & \multicolumn{1}{c|}{0.516}          & \multicolumn{1}{c|}{0.486}          & 0.455          \\ \cline{1-1} \cline{3-11} 
CNN-LSTM                &                                    & \multicolumn{1}{c|}{0.652}          & \multicolumn{1}{c|}{0.609}          & 0.367          & \multicolumn{1}{c|}{0.664}          & \multicolumn{1}{c|}{0.615}          & 0.359          & \multicolumn{1}{c|}{0.640}          & \multicolumn{1}{c|}{0.600}          & 0.379          \\ \cline{1-1} \cline{3-11} 
LipForensics          &                                    & \multicolumn{1}{c|}{\textbf{0.798}} & \multicolumn{1}{c|}{\textbf{0.774}} & \textbf{0.275} & \multicolumn{1}{c|}{\textbf{0.807}} & \multicolumn{1}{c|}{\textbf{0.785}} & \textbf{0.271} & \multicolumn{1}{c|}{\textbf{0.791}} & \multicolumn{1}{c|}{\textbf{0.763}} & \textbf{0.282} \\ \hline
EfficientNet V2-L      & \multirow{5}{*}{GBDF}              & \multicolumn{1}{c|}{0.843}          & \multicolumn{1}{c|}{0.825}          & 0.242          & \multicolumn{1}{c|}{0.849}          & \multicolumn{1}{c|}{0.833}          & 0.239          & \multicolumn{1}{c|}{0.837}          & \multicolumn{1}{c|}{0.817}          & 0.246          \\ \cline{1-1} \cline{3-11} 
XceptionNet            &                                    & \multicolumn{1}{c|}{0.847}          & \multicolumn{1}{c|}{0.825}          & 0.240          & \multicolumn{1}{c|}{0.854}          & \multicolumn{1}{c|}{0.834}          & 0.232          & \multicolumn{1}{c|}{0.840}          & \multicolumn{1}{c|}{0.816}          & 0.251          \\ \cline{1-1} \cline{3-11} 
MesoInception-4         &                                    & \multicolumn{1}{c|}{0.718}          & \multicolumn{1}{c|}{0.701}          & 0.324          & \multicolumn{1}{c|}{0.733}          & \multicolumn{1}{c|}{0.712}          & 0.309          & \multicolumn{1}{c|}{0.703}          & \multicolumn{1}{c|}{0.690}          & 0.340          \\ \cline{1-1} \cline{3-11} 
CNN-LSTM                &                                    & \multicolumn{1}{c|}{0.863}          & \multicolumn{1}{c|}{0.849}          & 0.225          & \multicolumn{1}{c|}{0.876}          & \multicolumn{1}{c|}{0.854}          & 0.211          & \multicolumn{1}{c|}{0.850}          & \multicolumn{1}{c|}{0.844}          & 0.235          \\ \cline{1-1} \cline{3-11} 
LipForensics          &                                    & \multicolumn{1}{c|}{\textbf{0.889}} & \multicolumn{1}{c|}{\textbf{0.866}} & \textbf{0.187} & \multicolumn{1}{c|}{\textbf{0.895}} & \multicolumn{1}{c|}{\textbf{0.878}} & \textbf{0.183} & \multicolumn{1}{c|}{\textbf{0.883}} & \multicolumn{1}{c|}{\textbf{0.854}} & \textbf{0.193} \\ \hline
\end{tabular}}
\end{center}
\end{table*}





\subsection{Performance differential of deepfake detectors on GBDF and DFDC-P test sets}

 
\begin{table*}[]
\caption {Evaluation of the DeepFake Detectors Across Males and Females when trained on FF++, GBDF and tested on \textbf{GBDF}. The metrics used are AUC, pAUC and EER. The performance differential~(P.D) is calculated as the absolute difference between EER of males and females.}
\label{Table5}
\begin{center}
\scalebox{0.78}{
\begin{tabular}{|c|c|ccc|ccc|ccc|c|}
\hline
\multirow{2}{*}{\textbf{Models}} & \multirow{2}{*}{\textbf{Training Dataset}} & \multicolumn{3}{c|}{\textbf{Overall}}                                                      & \multicolumn{3}{c|}{\textbf{Male}}                                                         & \multicolumn{3}{c|}{\textbf{Female}}                                                       & \multirow{2}{*}{\textbf{P.D}$\downarrow$} \\ \cline{3-11}
                                 &                                            & \multicolumn{1}{c|}{\textbf{AUC}}   & \multicolumn{1}{c|}{\textbf{pAUC}}  & \textbf{EER}   & \multicolumn{1}{c|}{\textbf{AUC}}   & \multicolumn{1}{c|}{\textbf{pAUC}}  & \textbf{EER}   & \multicolumn{1}{c|}{\textbf{AUC}}   & \multicolumn{1}{c|}{\textbf{pAUC}}  & \textbf{EER}   &                               \\ \hline
EfficientNet V2-L               & \multirow{5}{*}{FaceForensics++}                      & \multicolumn{1}{c|}{0.904}          & \multicolumn{1}{c|}{0.889}          & 0.179          & \multicolumn{1}{c|}{0.912}          & \multicolumn{1}{c|}{0.897}          & 0.171          & \multicolumn{1}{c|}{0.896}          & \multicolumn{1}{c|}{0.879}          & 0.187          & 0.016                         \\ \cline{1-1} \cline{3-12} 
XceptionNet                     &                                            & \multicolumn{1}{c|}{0.889}          & \multicolumn{1}{c|}{0.868}          & 0.217          & \multicolumn{1}{c|}{0.902}          & \multicolumn{1}{c|}{0.885}          & 0.206          & \multicolumn{1}{c|}{0.876}          & \multicolumn{1}{c|}{0.850}          & 0.228          & 0.022                         \\ \cline{1-1} \cline{3-12} 
MesoInception-4                  &                                            & \multicolumn{1}{c|}{0.769}          & \multicolumn{1}{c|}{0.747}          & 0.286          & \multicolumn{1}{c|}{0.759}          & \multicolumn{1}{c|}{0.742}          & 0.295          & \multicolumn{1}{c|}{0.779}          & \multicolumn{1}{c|}{0.750}          & 0.277          & 0.018                         \\ \cline{1-1} \cline{3-12} 
CNN-LSTM                         &                                            & \multicolumn{1}{c|}{0.909}          & \multicolumn{1}{c|}{0.888}          & 0.177          & \multicolumn{1}{c|}{0.917}          & \multicolumn{1}{c|}{0.898}          & 0.161          & \multicolumn{1}{c|}{0.901}          & \multicolumn{1}{c|}{0.877}          & 0.192          & 0.031                         \\ \cline{1-1} \cline{3-12} 
LipForensics                  &                                            & \multicolumn{1}{c|}{\textbf{0.942}} & \multicolumn{1}{c|}{\textbf{0.926}} & \textbf{0.109} & \multicolumn{1}{c|}{\textbf{0.938}} & \multicolumn{1}{c|}{\textbf{0.922}} & \textbf{0.113} & \multicolumn{1}{c|}{\textbf{0.947}} & \multicolumn{1}{c|}{\textbf{0.929}} & \textbf{0.105} & \textbf{0.008}                \\ \hline
EfficientNet V2-L               & \multirow{5}{*}{GBDF}                      & \multicolumn{1}{c|}{0.967}          & \multicolumn{1}{c|}{0.943}          & 0.052          & \multicolumn{1}{c|}{0.972}          & \multicolumn{1}{c|}{0.948}          & 0.050          & \multicolumn{1}{c|}{0.962}          & \multicolumn{1}{c|}{0.938}          & 0.054          & \textbf{0.004}                \\ \cline{1-1} \cline{3-12} 
XceptionNet                     &                                            & \multicolumn{1}{c|}{0.972}          & \multicolumn{1}{c|}{0.952}          & 0.046          & \multicolumn{1}{c|}{0.979}          & \multicolumn{1}{c|}{0.956}          & 0.043          & \multicolumn{1}{c|}{0.965}          & \multicolumn{1}{c|}{0.948}          & 0.049          & 0.006                         \\ \cline{1-1} \cline{3-12} 
MesoInception-4                  &                                            & \multicolumn{1}{c|}{0.819}          & \multicolumn{1}{c|}{0.800}          & 0.256          & \multicolumn{1}{c|}{0.828}          & \multicolumn{1}{c|}{0.805}          & 0.250          & \multicolumn{1}{c|}{0.811}          & \multicolumn{1}{c|}{0.795}          & 0.264          & 0.014                         \\ \cline{1-1} \cline{3-12} 
CNN-LSTM                         &                                            & \multicolumn{1}{c|}{0.975}          & \multicolumn{1}{c|}{0.957}          & 0.044          & \multicolumn{1}{c|}{0.983}          & \multicolumn{1}{c|}{0.964}          & 0.038          & \multicolumn{1}{c|}{0.967}          & \multicolumn{1}{c|}{0.950}          & 0.050          & 0.012                         \\ \cline{1-1} \cline{3-12} 
LipForensics                   &                                            & \multicolumn{1}{c|}{\textbf{0.978}} & \multicolumn{1}{c|}{\textbf{0.954}} & \textbf{0.039} & \multicolumn{1}{c|}{\textbf{0.982}} & \multicolumn{1}{c|}{\textbf{0.958}} & \textbf{0.036} & \multicolumn{1}{c|}{\textbf{0.974}} & \multicolumn{1}{c|}{\textbf{0.950}} & \textbf{0.042} & 0.006                         \\ \hline
\end{tabular}}
\end{center}
\end{table*}

Table ~\ref{Table5} shows the performance of the deepfake detectors across males and females when trained on FF++, GBDF, and tested on GBDF subject independent test set. Similarly, Table~\ref{Table6} shows the ACC, TPR, and FPR values associated with these models. The LipForensics model obtained the best results with an overall AUC of $0.978$, EER of $0.039$, and ACC of $0.967$ when trained and tested on GBDF.

When trained on FF++, the overall difference in the performance is $0.012$ and $0.0092$ in terms of pAUC and EER, respectively, across males and females. The overall difference in ACC, TPR and FPR is $0.010$, $0.0112$ and $0.012$, respectively, across males and females~(see~Table~\ref{Table6}). The least performance differential is obtained by EfficientNet V2-L when trained on FF++ and tested on GBDF.

\begin{table*}[]
\caption {ACC, TPR and FPR of the DeepFake Detectors Across Males and Females when trained on FF++, GBDF and tested on GBDF test set.} 
\label{Table6}
\begin{center}
\scalebox{0.88}{
\begin{tabular}{|c|c|ccc|ccc|ccc|}
\hline
\multirow{2}{*}{Models} & \multirow{2}{*}{Training Datasets} & \multicolumn{3}{c|}{Overall}                                                               & \multicolumn{3}{c|}{Male}                                                                  & \multicolumn{3}{c|}{Female}                                                                \\ \cline{3-11} 
                        &                                    & \multicolumn{1}{c|}{ACC}            & \multicolumn{1}{c|}{TPR}            & FPR            & \multicolumn{1}{c|}{ACC}            & \multicolumn{1}{c|}{TPR}            & FPR            & \multicolumn{1}{c|}{ACC}            & \multicolumn{1}{c|}{TPR}            & FPR            \\ \hline
EfficientNet V2-L      & \multirow{5}{*}{FaceForensics++}              & \multicolumn{1}{c|}{0.880}          & \multicolumn{1}{c|}{0.858}          & 0.205          & \multicolumn{1}{c|}{0.895}          & \multicolumn{1}{c|}{0.869}          & 0.194          & \multicolumn{1}{c|}{0.865}          & \multicolumn{1}{c|}{0.847}          & 0.220          \\ \cline{1-1} \cline{3-11} 
XceptionNet            &                                    & \multicolumn{1}{c|}{0.865}          & \multicolumn{1}{c|}{0.841}          & 0.222          & \multicolumn{1}{c|}{0.878}          & \multicolumn{1}{c|}{0.854}          & 0.209          & \multicolumn{1}{c|}{0.852}          & \multicolumn{1}{c|}{0.828}          & 0.235          \\ \cline{1-1} \cline{3-11} 
MesoInception-4         &                                    & \multicolumn{1}{c|}{0.745}          & \multicolumn{1}{c|}{0.721}          & 0.288          & \multicolumn{1}{c|}{0.735}          & \multicolumn{1}{c|}{0.715}          & 0.295          & \multicolumn{1}{c|}{0.754}          & \multicolumn{1}{c|}{0.727}          & 0.275          \\ \cline{1-1} \cline{3-11} 
CNN-LSTM                &                                    & \multicolumn{1}{c|}{0.883}          & \multicolumn{1}{c|}{0.868}          & 0.195          & \multicolumn{1}{c|}{0.894}          & \multicolumn{1}{c|}{0.884}          & 0.187          & \multicolumn{1}{c|}{0.872}          & \multicolumn{1}{c|}{0.854}          & 0.209          \\ \cline{1-1} \cline{3-11} 
LipForensics          &                                    & \multicolumn{1}{c|}{\textbf{0.925}} & \multicolumn{1}{c|}{\textbf{0.905}} & \textbf{0.178} & \multicolumn{1}{c|}{\textbf{0.920}} & \multicolumn{1}{c|}{\textbf{0.899}} & \textbf{0.180} & \multicolumn{1}{c|}{\textbf{0.930}} & \multicolumn{1}{c|}{\textbf{0.910}} & \textbf{0.175} \\ \hline
EfficientNet V2-L      & \multirow{5}{*}{GBDF}              & \multicolumn{1}{c|}{0.948}          & \multicolumn{1}{c|}{0.935}          & 0.154          & \multicolumn{1}{c|}{0.951}          & \multicolumn{1}{c|}{0.939}          & 0.149          & \multicolumn{1}{c|}{0.945}          & \multicolumn{1}{c|}{0.930}          & 0.159          \\ \cline{1-1} \cline{3-11} 
XceptionNet            &                                    & \multicolumn{1}{c|}{0.955}          & \multicolumn{1}{c|}{0.941}          & 0.144          & \multicolumn{1}{c|}{0.958}          & \multicolumn{1}{c|}{0.947}          & 0.139          & \multicolumn{1}{c|}{0.952}          & \multicolumn{1}{c|}{0.935}          & 0.147          \\ \cline{1-1} \cline{3-11} 
MesoInception-4         &                                    & \multicolumn{1}{c|}{0.802}          & \multicolumn{1}{c|}{0.778}          & 0.282          & \multicolumn{1}{c|}{0.808}          & \multicolumn{1}{c|}{0.788}          & 0.275          & \multicolumn{1}{c|}{0.796}          & \multicolumn{1}{c|}{0.770}          & 0.287          \\ \cline{1-1} \cline{3-11} 
CNN-LSTM                &                                    & \multicolumn{1}{c|}{0.953}          & \multicolumn{1}{c|}{0.939}          & 0.148          & \multicolumn{1}{c|}{0.959}          & \multicolumn{1}{c|}{0.941}          & 0.144          & \multicolumn{1}{c|}{0.949}          & \multicolumn{1}{c|}{0.935}          & 0.154          \\ \cline{1-1} \cline{3-11} 
LipForensics          &                                    & \multicolumn{1}{c|}{\textbf{0.967}} & \multicolumn{1}{c|}{\textbf{0.949}} & \textbf{0.142} & \multicolumn{1}{c|}{\textbf{0.971}} & \multicolumn{1}{c|}{\textbf{0.953}} & \textbf{0.135} & \multicolumn{1}{c|}{\textbf{0.965}} & \multicolumn{1}{c|}{\textbf{0.946}} & \textbf{0.145} \\ \hline
\end{tabular}}
\end{center}
\end{table*}


When trained on GBDF, the overall difference in the performance is $0.010$ and $0.008$ in terms of pAUC and EER, respectively, across males and females.
The overall difference in ACC, TPR and FPR is $0.008$,$0.010$ and $0.009$, respectively, across males and females~(see~Table~\ref{Table6}). The least performance differential is obtained by the LipForensics model when trained and tested on GBDF.  

Therefore, the difference in ACC, EER, TPR, and FPR decreased by $0.002$,
$0.001$, $0.0012$, and $0.003$, respectively, when using balanced GBDF as training and testing sets. Using balanced GBDF as a training and testing set, the highest bias mitigation is observed for CNN-LSTM and XceptionNet models. For CNN-LSTM, the difference in EER across gender reduced from $0.031$ to $0.012$ when trained with FF++ over GBDF as the training set (the test set is GBDF). Similarly, for XceptionNet, the difference in EER across gender reduced from $0.022$ to $0.006$ when trained with FF++ over GBDF as the training set (the test set is GBDF). Recall that the subjects do not overlap between the training and testing set of GBDF. Further, the samples in the training and testing set of GBDF are from three different deepfake datasets. 


\begin{table*}[]
\caption {Evaluation of the DeepFake Detectors Across Males and Females when trained on FF++, GBDF and tested on \textbf{DFDC-P}. The metrics used are AUC, pAUC and EER. The performance differential~(P.D) is calculated as the absolute difference between EER of males and females.}
\label{Table7}
\begin{center}
\scalebox{0.78}{
\begin{tabular}{|c|c|ccc|ccc|ccc|c|}
\hline
\multirow{2}{*}{\textbf{Models}} & \multirow{2}{*}{\textbf{Training Dataset}} & \multicolumn{3}{c|}{\textbf{Overall}}                                                      & \multicolumn{3}{c|}{\textbf{Male}}                                                         & \multicolumn{3}{c|}{\textbf{Female}}                                                       & \multirow{2}{*}{\textbf{P.D$\downarrow$}} \\ \cline{3-11}
                                 &                                            & \multicolumn{1}{c|}{\textbf{AUC}}   & \multicolumn{1}{c|}{\textbf{pAUC}}  & \textbf{EER}   & \multicolumn{1}{c|}{\textbf{AUC}}   & \multicolumn{1}{c|}{\textbf{pAUC}}  & \textbf{EER}   & \multicolumn{1}{c|}{\textbf{AUC}}   & \multicolumn{1}{c|}{\textbf{pAUC}}  & \textbf{EER}   &                               \\ \hline
EfficientNet V2-L               & \multirow{5}{*}{FaceForensics++}                      & \multicolumn{1}{c|}{0.659}          & \multicolumn{1}{c|}{0.634}          & 0.378          & \multicolumn{1}{c|}{0.665}          & \multicolumn{1}{c|}{0.641}          & 0.374          & \multicolumn{1}{c|}{0.653}          & \multicolumn{1}{c|}{0.626}          & 0.384          & \textbf{0.010}                \\ \cline{1-1} \cline{3-12} 
XceptionNet                     &                                            & \multicolumn{1}{c|}{0.642}          & \multicolumn{1}{c|}{0.624}          & 0.391          & \multicolumn{1}{c|}{0.649}          & \multicolumn{1}{c|}{0.632}          & 0.384          & \multicolumn{1}{c|}{0.635}          & \multicolumn{1}{c|}{0.617}          & 0.398          & 0.014                         \\ \cline{1-1} \cline{3-12} 
MesoInception-4                  &                                            & \multicolumn{1}{c|}{0.619}          & \multicolumn{1}{c|}{0.597}          & 0.421          & \multicolumn{1}{c|}{0.609}          & \multicolumn{1}{c|}{0.591}          & 0.432          & \multicolumn{1}{c|}{0.630}          & \multicolumn{1}{c|}{0.605}          & 0.410          & 0.022                         \\ \cline{1-1} \cline{3-12} 
CNN-LSTM                         &                                            & \multicolumn{1}{c|}{0.667}          & \multicolumn{1}{c|}{0.648}          & 0.372          & \multicolumn{1}{c|}{0.675}          & \multicolumn{1}{c|}{0.661}          & 0.356          & \multicolumn{1}{c|}{0.659}          & \multicolumn{1}{c|}{0.635}          & 0.382          & 0.026                         \\ \cline{1-1} \cline{3-12} 
LipForensics                   &                                            & \multicolumn{1}{c|}{\textbf{0.718}} & \multicolumn{1}{c|}{\textbf{0.705}} & \textbf{0.312} & \multicolumn{1}{c|}{\textbf{0.724}} & \multicolumn{1}{c|}{\textbf{0.710}} & \textbf{0.306} & \multicolumn{1}{c|}{\textbf{0.712}} & \multicolumn{1}{c|}{\textbf{0.701}} & \textbf{0.319} & 0.013                         \\ \hline
EfficientNet V2-L               & \multirow{5}{*}{GBDF}                      & \multicolumn{1}{c|}{0.684}          & \multicolumn{1}{c|}{0.662}          & 0.349          & \multicolumn{1}{c|}{0.689}          & \multicolumn{1}{c|}{0.665}          & 0.345          & \multicolumn{1}{c|}{0.680}          & \multicolumn{1}{c|}{0.661}          & 0.354          & \textbf{0.009}                \\ \cline{1-1} \cline{3-12} 
XceptionNet                     &                                            & \multicolumn{1}{c|}{0.668}          & \multicolumn{1}{c|}{0.652}          & 0.368          & \multicolumn{1}{c|}{0.675}          & \multicolumn{1}{c|}{0.657}          & 0.363          & \multicolumn{1}{c|}{0.663}          & \multicolumn{1}{c|}{0.649}          & 0.374          & 0.011                         \\ \cline{1-1} \cline{3-12} 
MesoInception-4                  &                                            & \multicolumn{1}{c|}{0.615}          & \multicolumn{1}{c|}{0.592}          & 0.427          & \multicolumn{1}{c|}{0.621}          & \multicolumn{1}{c|}{0.596}          & 0.419          & \multicolumn{1}{c|}{0.608}          & \multicolumn{1}{c|}{0.585}          & 0.432          & 0.013                         \\ \cline{1-1} \cline{3-12} 
CNN-LSTM                         &                                            & \multicolumn{1}{c|}{0.689}          & \multicolumn{1}{c|}{0.665}          & 0.343          & \multicolumn{1}{c|}{0.683}          & \multicolumn{1}{c|}{0.658}          & 0.350          & \multicolumn{1}{c|}{0.694}          & \multicolumn{1}{c|}{0.674}          & 0.334          & 0.016                         \\ \cline{1-1} \cline{3-12} 
LipForensics                   &                                            & \multicolumn{1}{c|}{\textbf{0.732}} & \multicolumn{1}{c|}{\textbf{0.721}} & \textbf{0.299} & \multicolumn{1}{c|}{\textbf{0.736}} & \multicolumn{1}{c|}{\textbf{0.724}} & \textbf{0.292} & \multicolumn{1}{c|}{\textbf{0.727}} & \multicolumn{1}{c|}{\textbf{0.716}} & \textbf{0.307} & 0.015                         \\ \hline
\end{tabular}}
\end{center}
\end{table*}

Table ~\ref{Table7} shows the performance of the deepfake detectors across males and females when trained on FF++, GBDF, and tested on DFDC-P. \textbf{Note that DFDC-P dataset has not been used in the creation of the GBDF dataset}. As the original DFDC-P test set does not contain subject IDs, the subset of DFDC training set is manually annotated with gender labels and used as a test set for this study.
Overall, low performance is obtained for all the models on DFDC dataset. This is because DFDC consist of low quality videos that are diverse across gender, skin-tone and age-group.
The LipForensics model obtained the best results with an overall AUC of $0.732$, EER of $0.299$ when tested on DFDC-P.

When trained on FF++, the overall difference in the performance is $0.010$ and $0.0082$ in terms of pAUC and EER, respectively, across males and females. The least performance differential is obtained by EfficientNet V2-L when trained on FF++ and tested on DFDC-P.

When trained on GBDF, the overall difference in the performance is $0.004$ and $0.007$ in terms of pAUC and EER, respectively, across males and females. The least performance differential is obtained by EfficientNet V2-L when trained on GBDF and tested on DFDC-P. 
\textit{These results suggest that using our gender-balanced GBDF training set, bias is mitigated across gender even on an external DFDC-P dataset, not used in the creation of GBDF.}

\begin{figure*}[htbp]
\centerline{\includegraphics[width=1.0\textwidth]{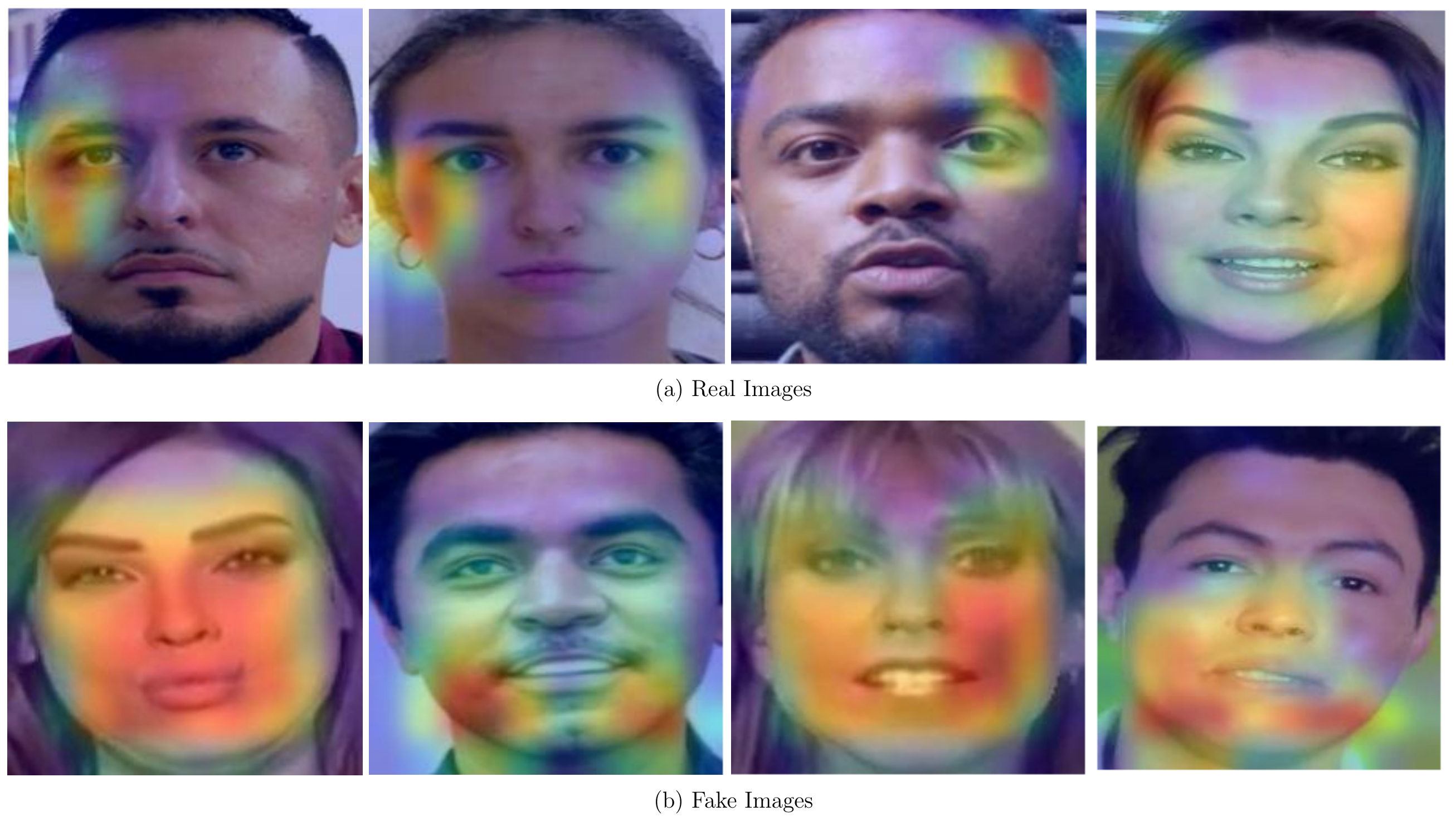}}
\caption{Grad-CAM visualization of the EfficientNet V2-L based deepfake detector on randomly selected live and fake samples from males and females. The distinctive image regions used by the CNN model for deepfake detection differs across gender.} 
\label{fig3}
\end{figure*}

Finally, we also used Explainable AI (XAI) based Gradient weighted
Class Activation Mapping (Grad-CAM)~\cite{8237336} visualization to understand the distinctive image regions used by the CNN models in detecting deepfakes across gender. GRAD-CAM uses the gradients of any target concept to generate a coarse localization map that highlights distinctive image regions used for making a decision/prediction~\cite{8237336}. Figure~\ref{fig3} shows the GRAD-CAM visualization of the EfficientNet V2-L-based deepfake detector for live and fake images for males and females. This detector was trained on GBDF dataset. The highly activated region is shown by the red zone on the map, followed by green and blue zones. It can be seen that the highly activated region is the cheek for females and the ocular region for males. For fake images, the mouth and cheek region for males and the complete face region for females are the most activated region. These results were consistent across the datasets depending on the deepfake generation technique. Therefore, different image regions were used by the deepfake detector for live and fake classification across gender.





In \textbf{summary}, males outperformed females for most of the models, with the disparity of about $0.034$ in terms of EER in the range $[0, 1]$ for MesoInception-4 model. The shallow MesoInception-4 model demonstrated high performance differential across gender for most of the experiments. The LipForensics model, on the other hand, obtained least disparity across gender for most of the experiments. This is because it uses mouth crops for mouth motion analysis. Thus the impact of gendered differences in facial images attributed to bias are mitigated to a major extent. 
When trained on FF+, males outperformed females for the majority of the models despite having a lower percentage than females. As large number of the videos in FaceForensics++ are irregular deepfakes, it is not certain which gender-group-related features are dominant in irregular facial swaps.
The gender-balanced GBDF training set reduced the performance difference over FF++, with the highest being from $0.031$ to $0.012$ in terms of EER across males and females when tested on GBDF test set. The advantage of using GBDF training set towards gender fair deepfake classification is also noticed for an external DFDC-P set. The grad-CAM visualization suggests the distinctive image regions used by the CNN model for deepfake classification differs across gender. As these automated deepfake detection systems are used at the mass-level for audit of the social media data, even a small reduction in the bias across demographics would positively impact millions of people belonging to the deprived sub-group. 

\section{Conclusion and Future Research Directions}
With the volume of deepfake videos showing staggering growth, there is a growing reliance on automated systems to combat deepfakes. For the massive rollout of this high-impact technology, it becomes vital to understand all the societal aspects including demographic disparities. In this work, we thoroughly examined the fairness of the deepfake detectors on gender-aware deepfake datasets. On manual annotation of gender labels, we found that current deepfake datasets have a highly skewed distribution across gender and contain irregular swaps. The popular deepfake detectors have exhibited disparities in the performance across gender when evaluated on gender-aware datasets, with mostly males outperforming females. This suggest an additional threat imposed by deep fake technology on female subjects, primarily due to the performance differential of SOTA deep fake detectors. 

However, using our gender-balanced GBDF dataset, the unequal performance of the deepfake detectors across gender is mitigated to some extent. Our work echoes the importance of benchmarking demographically balanced and labeled deepfake datasets to facilitate intersectional subgroup-based audits of existing deepfake detectors along with the cause and effect analysis. As a part of future work, fairness of the deepfake detectors will also be evaluated across race. Further, the fairness-aware deepfake detectors will be developed for increased demographic transparency and accountability of these high-impact systems.

\section{Acknowledgement}
This work is supported in part from National Science Foundation~(NSF) award no. $2129173$.   
The research infrastructure used in this study is supported in part from a grant no. $13106715$ from
the Defense University Research Instrumentation Program (DURIP) from Air Force Office of Scientific Research.
%
%
%
%
%
 \bibliographystyle{splncs04}
\bibliography{GBDF_2022}
%
\end{document}